\documentclass{l4dc2025}
\usepackage{amssymb}
\usepackage{amsmath}
\usepackage{breqn}
\usepackage{bbm}
\usepackage{color, soul}
\usepackage{amsfonts}
\usepackage{latexsym}
\usepackage{graphicx}
\usepackage{multirow} 
\usepackage{booktabs}
\usepackage{algorithm,algpseudocode}

\def\figpath{./figs/}

\title[LLM-Enhanced Path Planning]{LLM-Enhanced Path Planning: Safe and Efficient Autonomous Navigation with Instructional Inputs}
\usepackage{times}

\author{%
 \Name{Pranav Doma} \Email{pd2365@nyu.edu }\\
 \addr Department of Mechanical and Aerospace Engineering, Tandon School of Engineering, New York University, 6 Metro Tech, Brooklyn, NY, USA 1
 \AND
 \Name{Aliasghar {Arab}} \Email{aliasghar.arab@nyu.edu}\\
 \addr Department of Mechanical and Aerospace Engineering, Tandon School of Engineering, New York University, 6 Metro Tech, Brooklyn, NY, USA 1%
 \AND
 \Name{Xuesu {Xiao}} \Email{xiao@gmu.edu}\\
 \addr Department of Computer Science, George Mason University 2%
}

\begin{document}
\maketitle

\begin{abstract}%
Autonomous navigation guided by natural language instructions is essential for improving human-robot interaction and enabling complex operations in dynamic environments. While large language models (LLMs) are not inherently designed for planning, they can significantly enhance planning efficiency by providing guidance and informing constraints to ensure safety. This paper introduces a planning framework that integrates LLMs with 2D occupancy grid maps and natural language commands to improve spatial reasoning and task execution in resource-limited settings. By decomposing high-level commands and real-time environmental data, the system generates structured navigation plans for pick-and-place tasks, including obstacle avoidance, goal prioritization, and adaptive behaviors. The framework dynamically recalculates paths to address environmental changes and aligns with implicit social norms for seamless human-robot interaction. Our results demonstrates the potential of LLMs to design context-aware system to enhance navigation efficiency and safety in industrial and dynamic environments.
\end{abstract}
\begin{keywords}%
Large Language Models, Path Planning, Safe Autonomous Navigation.
\end{keywords}

\section{Introduction}
Autonomous robotic navigation has seen substantial progress through the integration of multimodal data, LLMs, and occupancy grid representations. While traditional path planning algorithms like Rapidly-exploring Random Trees (RRT*) and A* excel in static environments, they struggle with adaptability and contextual reasoning in dynamic and uncertain scenarios. These methods lack the capability for semantic understanding or real-time, instruction-based adjustments, limiting their utility in environments with abrupt changes, such as navigating around potholes~\citep{li2023robochat} or through repair zones~\citep{an2023bevbert, dorbala2022clip}.

\begin{figure}[t!]
\centering
	\includegraphics[width=3.7in]{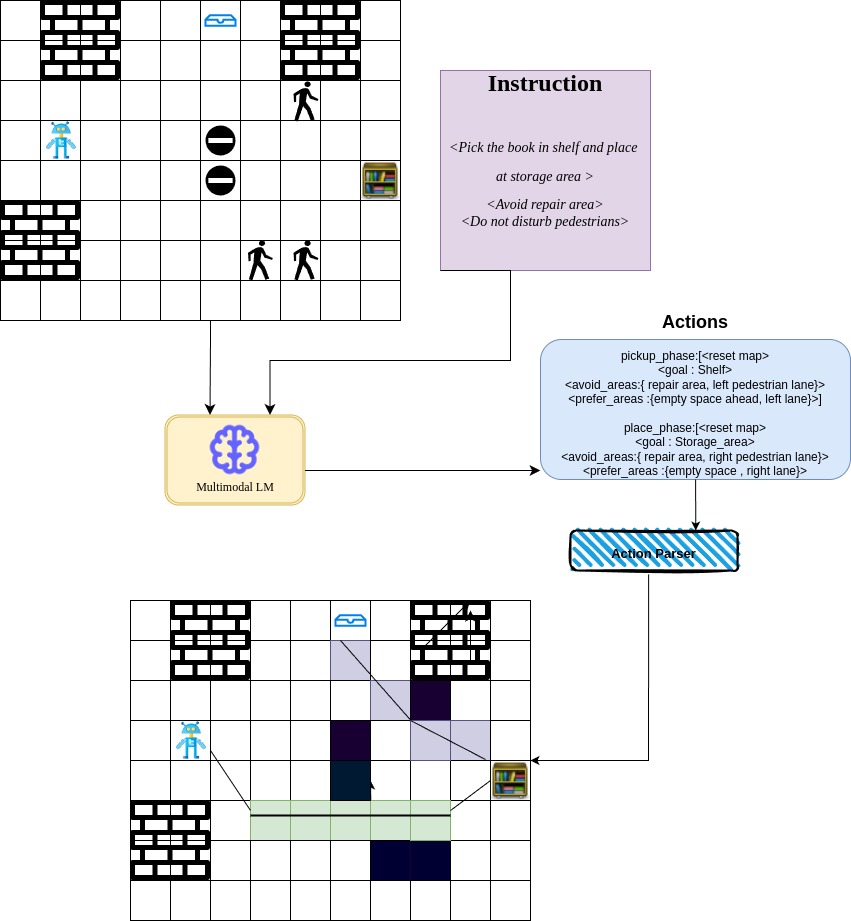}
	\caption{High-level diagram for our proposed instructed planning method.}
	\label{fig:diagram}
	\vspace{-3mm}
\end{figure}

\noindent Recent methods like BEVBert~\citep{an2023bevbert}, CLIP-NAV~\citep{dorbala2022clip}, and InstructNav~\citep{long2024instructnav} showcase the potential of multimodal inputs and LLMs for intuitive navigation but often depend on high-fidelity sensors and comprehensive maps, limiting their applicability in environments with unreliable sensors or unexpected obstacles. To overcome these challenges, we introduce the Dynamic Chain of Instruction-based Planning (DCIP) framework, which combines instruction-based planning with dynamic occupancy grid updates. DCIP enables robots to interpret high-level instructions via natural language understanding (NLU) and adapt navigation strategies in real-time. For example, detecting a pothole triggers DCIP to classify the area as avoidable, ensuring safe, efficient replanning with minimal computational overhead. By integrating this approach with lightweight planners like Model Predictive Control (MPC) and Dynamic Window Approach (DWA)~\citep{song2023llm, lin2024navcot}, DCIP enhances spatial awareness, safety, and behavior customization. The main contributions for our instructed planning method shown in Figure~\ref{fig:diagram}, are listed as
\begin{itemize}
    \item \textbf{Enhanced Environmental Understanding via Multimodal Integration}: DCIP combines visual and language inputs to semantically interpret complex scenarios, such as avoiding obstacles like potholes or repair zones, beyond the capabilities of traditional 2D LiDAR systems.

    \item \textbf{Real-Time Instruction-Based Planning}: By leveraging NLU, DCIP dynamically adjusts navigation strategies and updates occupancy grids in response to environmental changes, enabling flexible and adaptive robot behavior.

    \item \textbf{Dynamic Occupancy Grid Manipulation}: DCIP employs semantic landmarks to update occupancy grids in real-time, ensuring efficient replanning and enhanced spatial awareness, such as classifying newly detected obstacles for safe navigation.
\end{itemize}

\begin{figure}[t!]
\centering
	\includegraphics[width=3.3in]{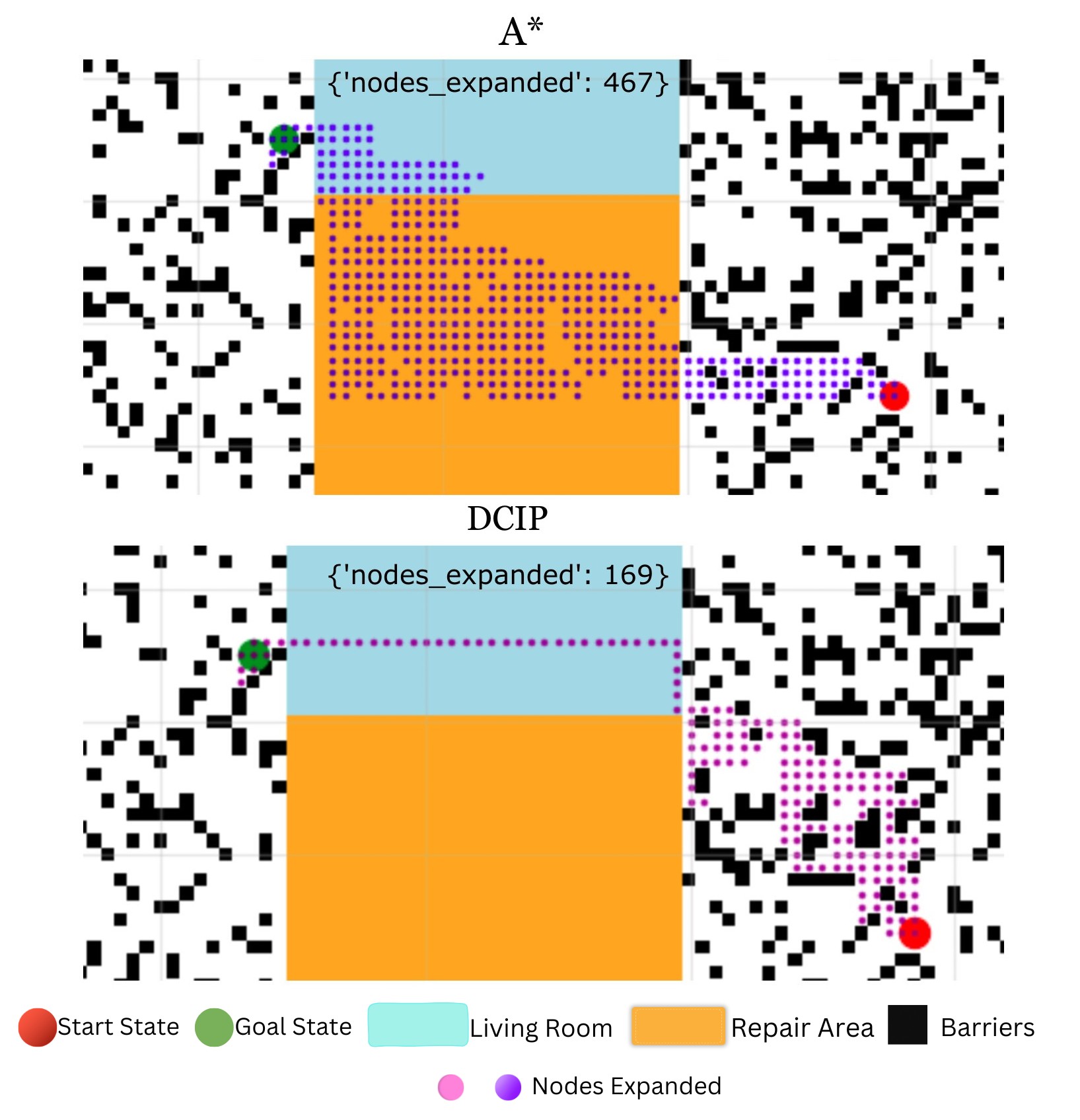}
	\caption{Comparison of the Instruction-Based Planning Strategies on a Grid Using the Llama3.1 Model. \textbf{blue}: traditional \textbf{A star}; \textbf{purple}: proposed \textbf{DCIP }algorithm.}
	\label{fig:learning1}
\end{figure}

\noindent Our proposed DCIP method bridges the gap between traditional static path planning and dynamic environmental adaptation by combining the robustness of occupancy grids with the flexibility of instruction-based planning. In Figure~\ref{fig:learning1}, paths generated by the traditional A* algorithm (blue) and the DCIP algorithm (purple) are compared under three instruction sets, highlighting DCIP's adaptability to prioritize speed, reduced turns, and safety. The middle panel shows a path minimizing turns, which compromises safety and obstacle avoidance. The right panel highlights a secure navigation path, emphasizing safety, obstacle avoidance, and repair area detours. These comparisons demonstrate the model's adaptability to different strategic goals and instruction adherence. Our proposed approach enhances task reliability and ensures safer navigation in complex, unpredictable environments, marking a significant step forward in autonomous robotic navigation.  

\section{Literature Review}
\label{sec:literature_review}

Robotic navigation has advanced considerably with the integration of multimodal data, occupancy grids, and instruction-based planning. This section reviews key research in these domains, examining their achievements and limitations. The DCIP framework extends these foundations to overcome existing challenges and improve autonomous navigation performance.

\subsection{Multimodal Navigation}
Multimodal approaches have significantly enhanced robotic navigation by integrating diverse sensory inputs for richer environmental understanding and more effective task execution~\citep{panigrahi2023study}. BEVBert~\citep{an2023bevbert} employs multimodal map pre-training for language-guided navigation, enabling robots to act on natural language instructions within spatial contexts. Similarly, CLIP-NAV~\citep{dorbala2022clip} leverages the CLIP model for zero-shot vision-and-language navigation, allowing robots to follow textual instructions without task-specific training. L3mvn~\citep{yu2023l3mvn} advances this field by using LLMs to interpret complex visual instructions for target-based navigation, while \cite{hong2023learning} embed semantic information into occupancy grids through supervised learning, enhancing perceptual capabilities. Despite these strides, most existing frameworks rely on high-fidelity sensor data and detailed environmental models, which limit their effectiveness in dynamic or sensor-degraded settings. The proposed DCIP framework addresses these challenges by using occupancy grids as a central component, ensuring robust and adaptable navigation even under compromised sensor conditions.

\subsection{Occupancy Grid Utilization}
Occupancy grids are a cornerstone of robotic navigation, offering a probabilistic representation of the environment by categorizing cells as occupied, free, or unknown~\citep{schreiber2021dynamic}. Extensions like SCOPE~\citep{xie2024scope} incorporate uncertainty modeling for safer navigation in dynamic settings, while dynamic grid mapping with recurrent neural networks~\citep{schreiber2021dynamic} enables real-time updates for adaptive responses. Semantic map supervision~\citep{hong2023learning} further enriches grids with semantic information for more informed decision-making. Beyond traditional grids, approaches such as by \cite{krantz2020beyond} integrate vision and language navigation in continuous environments, moving past static nav-graphs. While effective, these methods often struggle with rapidly changing conditions. The proposed DCIP framework addresses these limitations by enabling dynamic manipulation and real-time updates of occupancy grids, ensuring adaptability to environmental changes and user instructions.

\subsection{Instruction-Based Planning and Large Language Models}
The integration of LLMs has transformed instruction-based planning in robotic navigation. Systems like InstructNav~\citep{long2024instructnav} and LLM-Planner~\citep{song2023llm} enable robots to interpret and execute complex instructions in unexplored environments using zero-shot and few-shot learning techniques. Plan-and-Solve Prompting~\citep{wang2023plan} enhances chain-of-thought reasoning for decomposing high-level tasks into actionable sub-tasks, while NavCoT~\citep{lin2024navcot} refines vision-and-language navigation with disentangled reasoning processes for greater reliability. Despite these advancements, these models often operate within static planning frameworks, where action sequences are predetermined based on initial observations~\citep{arab2021instructed}. Research such as Controllable Navigation Instruction Generation~\citep{kong2024controllable} and Esc~\citep{zhou2023esc} highlights how chain-of-thought prompting and commonsense constraints can improve navigation in zero-shot scenarios~\citep{wei2022chain}. The proposed DCIP framework stands apart by adopting a dynamic planning approach, continuously updating action sequences in response to real-time observations and user interactions. By integrating instruction-based planning with dynamic occupancy grid manipulation, DCIP ensures that navigation plans adapt to unexpected obstacles and degraded sensor data, maintaining accuracy and alignment with the environment.

\noindent The reviewed literature highlights the transformative impact of integrating multimodal data, occupancy grids, and large language models in robotic navigation. However, challenges persist in achieving robust adaptability and safety in dynamic, uncertain environments. The DCIP framework bridges these gaps by combining dynamic occupancy grid updates with real-time instruction-based planning, enhancing spatial awareness, adaptability, and safety. This advancement positions DCIP as a key innovation toward more reliable and efficient autonomous navigation systems.

\section{Problem Statement}
\label{sec:problem_statement}

Autonomous robotic navigation in dynamic and complex environments necessitates the ability to interpret and act upon high-level instructions effectively. Traditional path planning algorithms, such as Rapidly-exploring Random Trees (RRT*)~\citep{noreen2016optimal} and A*~\citep{hart1968formal}, excel in static scenarios but often falter in the face of dynamic obstacles and varying task objectives. These methods typically lack the semantic understanding and real-time adaptability required for nuanced navigational tasks driven by natural language instructions. This research addresses the challenge of Instruction-Based Navigation, where a robot must navigate from a starting position to a goal location based on natural language instructions, while dynamically adapting to environmental changes. The environment is modeled as a two-dimensional occupancy grid $E$, where each cell $E(x, y)$ represents the occupancy state:
\begin{equation}
E(x, y) \in \{0, 1\}
\end{equation}

\noindent where, $0$ represents \textit{Free Space} and
$1$ represents \textit{Occupied Space}. At each discrete time step $t$, the robot receives observations:
\begin{equation}
O_t = \{G_t, P_t, S_t\}
\end{equation}
\noindent where, $G_t$ is current 2D occupancy grid and $P_t = (x_t, y_t, \theta_t)$ is a vector repressing robot’s pose, comprising position $[x_t, y_t]$ and orientation $\theta_t$, and $S_t$ is a set of detected semantic landmarks within the environment. Given a high-level navigation instruction $I$ in natural language, the primary objective is to generate a sequence of actions $A_t = \{a_1, a_2, \ldots, a_n\}$ that transitions the robot from its current state to a desired goal state $G$. This sequence must satisfy
\begin{itemize}
    \item \textbf{Semantic Alignment}: Actions must adhere to the semantic intent of $I$, ensuring contextual relevance.
    \item \textbf{Obstacle Avoidance}: Navigate around occupied spaces and dynamically detected obstacles to prevent collisions.
    \item \textbf{Dynamic Adaptability}: Adjust the navigational strategy in real-time in response to environmental changes.
\end{itemize}

To achieve these objectives, the proposed DCIP framework leverages NLU to interpret instructions, integrates semantic landmark data for informed decision-making, and manipulates the occupancy grid by assigning traversal costs to preferred and avoidable areas. This integration facilitates the selection of optimal navigation paths that are both efficient and contextually appropriate, ensuring reliable task completion in diverse and unpredictable environments.

\section{Method Overview}
\label{sec:method_overview}

The DCIP framework is engineered to enable autonomous robots to execute complex navigational tasks based on natural language instructions. By integrating NLU, multimodal understanding, and dynamic occupancy grid manipulation, DCIP offers a robust solution for instruction-driven navigation. This section outlines the core components and operational workflow of DCIP, emphasizing its mathematical foundation and modular design. The DCIP algorithm in ~\ref{alg:DCIP} is designed to interpret high-level instructions and dynamically generate a sequence of actions for autonomous navigation. By processing initial observations and continuously adapting to real-time environmental changes, DCIP ensures efficient and goal-oriented pathfinding. The algorithm flexibly handles various instructions, determining appropriate goals and actions based on the given directives.

\begin{algorithm}[h!]
\caption{Dynamic Chain of Instruction-based Planning}
\label{alg:DCIP}
\textbf{Input:} High-level instruction $I$, initial observation $O_0$\;
\textbf{Output:} Sequence of actions $A_t$\;
Initialize DCIP with instruction $I$ and observation $O_0$\;
Parse instruction $I$ to identify goals and constraints\;
Generate initial action sequence $A_0$ based on $I$ and $O_0$\;
\textbf{Execute Initial Actions}:
\For{each action $a \in A_0$}
{
Execute action $a$\;
Update occupancy grid based on action $a$\;
}
\textbf{Real-time Adaptation Loop}:
\While{task not completed}{
Capture current observation $O_t$\;
Update goals and constraints based on $I$ and $O_t$\;
Generate new action sequence $A_t$ based on updated goals\;
   \For{each action $a \in A_t$}
   {
Execute action $a$\;
Update occupancy grid based on action $a$\;
   }
}
\textbf{End of Algorithm}
\end{algorithm}

\subsection{Instruction Dissection}
The DCIP framework begins by processing the natural language instruction $I$ through an NLU module to extract critical components necessary for navigation:

\begin{equation}
I \rightarrow \{ \text{Task Definition}, \text{Action Format}, \text{Semantic Constraints}\}
\end{equation}

\begin{itemize}
    \item \textbf{Task Definition}: Identifies the primary objective (e.g., ``navigate to Shelf 3'').
    \item \textbf{Action Format}: Defines the structure of actionable commands (e.g., RESET\_MAP, AVOID\_AREAS).
    \item \textbf{Semantic Constraints}: Extracts specific conditions or preferences (e.g., ``avoid repair area'', ``prefer open lanes'').
\end{itemize}

\subsection{Action Format and Prediction}
Within the DCIP framework, actions are systematically categorized to modify the occupancy grid $G_t$ or to set navigation goals $G_{\text{goal}}$. These actions enable dynamic adaptation to varying instructions and environmental conditions. Formally, let $\mathcal{A}$ represent the set of possible actions, where each action $a \in \mathcal{A}$ is associated with a transformation function $f_a$. Hence, \textit{Map Resetting}, resets the occupancy grid $G_t$ to its initial state $G_0$ as

\begin{equation}
    G' = f_{\text{reset}}(G_t) = G_0,
\end{equation}
\noindent In additions, \textit{Cost Modification} adjusts traversal costs within specified regions $R$ based on semantic constraints as
\begin{equation}
    G'(x, y) =
\begin{cases}
    G_t(x, y), & \text{if } (x, y) \notin R, \\
    C_{\text{modify}}, & \text{if } (x, y) \in R,
\end{cases}
\end{equation}
\noindent where, $C_{\text{modify}}$ represents the cost adjustment factor. Eventually, \textit{Goal Setting} establishes the target location for navigation as
\begin{equation}
    G_{\text{goal}} = (x_{\text{goal}}, y_{\text{goal}}).
\end{equation}

\subsection{General Action Framework}
Each action $a \in \mathcal{A}$ is defined as a function $f_a$ that transforms the current state of the occupancy grid $G_t$ or sets new navigation goals $G_{\text{goal}}$. This generalized approach ensures the framework's adaptability across diverse environments and instructions.

\begin{equation}
\mathcal{A} = \left\{ a \mid a = f_a(G_t, \theta_a) \right\},
\end{equation}

\noindent where $\theta_a$ denotes parameters specific to action $a$, such as regions $R$ or goal coordinates $G_{\text{goal}}$. The execution of actions within the DCIP framework can be formalized as follows:
\begin{align}
    & \text{Given: } G_t, \quad I \quad (\text{Instruction}), \\ \nonumber
    & \text{Parse } I \rightarrow A, \quad \text{where } A \subseteq \mathcal{A}, \\ \nonumber
    & \text{For each } a \in A, \quad G' = f_a(G_t, \theta_a), \\ \nonumber
    & \text{Update } G_t \leftarrow G' \quad G_{goal} \leftarrow G'_{goal}.
\end{align}
\noindent where $I$ is the natural language instruction, $A$ is the set of actions derived from parsing $I$, and $\theta_a$ are the parameters associated with action $a$. As an example, consider an instruction $I$ to ``navigate to the target while avoiding obstacles.'' However, Goal is specified on the occupancy grid, so updating the occupancy grid also includes updating the goal which is not always the case. The NLU module extracts the relevant actions $A = \{ \text{SET\_GOAL}, \text{MODIFY\_COST} \}$. The framework executes these actions as follows:

SET\_GOAL: Establishes the updated target location:
   \begin{equation}
   G'_{\text{goal}} = (x'_{\text{target}}, y'_{\text{target}}).
   \end{equation}

MODIFY\_COST: Assigns higher traversal costs to regions identified as obstacles:
\begin{equation}
   G'(x, y) =
   \begin{cases}
       G_t(x, y), & \text{if } (x, y) \notin \text{Obstacles}, \\
       C_{\text{high}}, & \text{if } (x, y) \in \text{Obstacles}.
   \end{cases}
\end{equation}
   
\noindent This generalized structure ensures that DCIP remains adaptable to various instructions and environmental dynamics, providing robust and context-aware navigation capabilities.

% \subsection{Task Region Selection}
% DCIP dynamically selects sub-goal regions and categorizes areas into avoidable and preferred based on the parsed instruction. This involves:
% \begin{itemize}
%     \item \textbf{Sub-Goal Identification}: Determines intermediate targets to facilitate path planning.
%     \item \textbf{Region Classification}: Utilizes semantic landmarks to classify regions within the occupancy grid.
% \end{itemize}

\subsection{Multimodal Understanding and Action Parsing}
DCIP integrates multimodal data (e.g., semantic landmarks from visual inputs) and parses actions to update the occupancy grid accordingly:
\begin{equation}
\text{Multimodal Data} + \text{Parsed Actions} \rightarrow \text{Updated Occupancy Grid}
\end{equation}
This integration ensures that the robot can interpret complex instructions and translate them into spatial modifications within the environment.

\section{Experiments and Results}
To evaluate the effectiveness of the DCIP framework, we conducted a series of experiments using mathematical simulations and  a Husky-UR3 compound robot within an industrial simulation environment in Gazebo. The primary objective was to assess DCIP’s ability to interpret and execute complex natural language instructions for pick-and-place tasks while adhering to specified constraints such as avoiding restricted areas and maintaining safe distances from pedestrians.

\subsection{Performance Comparison}
We evaluated the performance of three language models—Mistral, Llama3.1, and Llama3—across four navigation strategies. The evaluation focused on three key metrics to assess model performance. \textit{Instruction Adherence} measured how accurately the model's outputs aligned with the provided navigation strategy instructions. \textit{Completeness of Response} evaluated whether the generated outputs included all necessary components without omissions, ensuring task fulfillment. Finally, \textit{Safety Compliance} assessed the system's ability to avoid restricted or socially unsafe areas, emphasizing adherence to safety guidelines during navigation. These metrics collectively provided a comprehensive assessment of the model's effectiveness in interpreting and executing natural language instructions for autonomous navigation.

\begin{table}[ht]
\centering
\caption{Model Performance Comparison Across Strategies}
\label{table:model_performance}
\scriptsize
\begin{tabular}{|l|p{3.7cm}|p{1.9cm}|p{2.05cm}|p{1.5cm}|}
\hline
\textbf{Model} & \textbf{Strategy} & \textbf{Instruction Adherence (\%)} & \textbf{Completeness of Response (\%)} & \textbf{Safety Compliance (\%)} \\ \hline
\multirow{3}{*}{Mistral} & Navigate Quickly & 28.57 & 42.85 & 14.28 \\ 
 & Maximize Safety & 85.71 & 71.42 & 89.28 \\ 
 & Balance Efficiency and Safety & 71.42 & 85.71 & 89.28 \\ \hline
\multirow{3}{*}{Llama3.1} & Navigate Quickly & 57.14 & 71.42 & 64.28 \\
 & Maximize Safety & 85.71 & 71.42 & 91.42 \\ 
 & Balance Efficiency and Safety & 89.28 & 88.57  & 92.85 \\ \hline
\multirow{3}{*}{LLama3} & Navigate Quickly & 82.14 & 64.28 & 71.42  \\ 
 & Maximize Safety & 89.28 & 85.71  & 92.85   \\ 
 & Balance Efficiency and Safety & 92.85 & 96.42 & 96.42 \\ \hline
\end{tabular}
\end{table}

\noindent The analysis from Table~\ref{table:model_performance} reveals that Llama3 consistently outperformed Mistral and Llama3.1 across all evaluated strategies, excelling in Instruction Adherence, Completeness, and Compliance. While Mistral demonstrated moderate performance, it showed particular strength in the ``Navigate Quickly'' strategy, reflecting its capability to prioritize speed and efficiency. In contrast, Llama3.1 lagged behind the other models, especially in maintaining relevance to the intended navigation strategy, indicating a need for improved alignment between instructions and execution. These results underscore Llama3's superior ability to interpret and execute complex navigation tasks effectively.

\subsubsection{DCIP Core Performance}
In addition to model-specific evaluations, we assessed DCIP's core algorithmic performance metrics--Nodes Expanded, Search Time, Path Cost, Path Length, and Number of Turns to understand its efficiency and effectiveness in planning navigation paths.

The performance evaluation encompasses five key metrics:
\begin{itemize}
    \item Nodes Expanded: Counts unique positions explored
        \begin{equation}
        N_{expanded} = |\{n \in \text{Nodes} : n \text{ processed}\}|,
        \end{equation}

    \item Search Time: Execution duration
        \begin{equation}
        T_{search} = T_{end} - T_{start}
        \end{equation}
    
    \item Path Cost: Sum of node costs with zone weights
        \begin{equation}
        C_{path} = \sum_{i=1}^{n} (g_i + \text{zone\_cost}_i),
        \end{equation}
    
    \item Path Length: Total steps in path
        \begin{equation}
        L_{path} = |P| - 1,
        \end{equation}

    \item Number of Turns: Direction changes
        \begin{equation}
        N_{turns} = \sum_{i=1}^{n-2} \mathbb{1}_{v_i \neq v_{i+1}}
        \end{equation}
\end{itemize}

The performance data reveals distinct behavioral patterns across different models and strategies:

\begin{itemize}
    \item Computational Efficiency: While A* maintained consistent exploration patterns, DCIP demonstrated adaptive behavior. Llama models showed remarkable efficiency in quick navigation scenarios, exploring significantly fewer nodes than Mistral.

    \item Search Time Correlation: Search duration typically aligned with exploration patterns, with quick navigation strategies showing notably faster completion times in Llama models.

    \item Path Cost Variations: Mistral's quick navigation strategy revealed an interesting trade-off, where it prioritized speed over zone preferences, resulting in higher path costs but faster navigation through typically avoided areas.

    \item Path Length Consistency: Most DCIP implementations maintained optimal path lengths comparable to A*, suggesting effective route planning despite varying constraints.

    \item Turn Optimization: DCIP consistently generated smoother paths with fewer direction changes compared to A*, with particularly notable improvements in quick navigation scenarios.
\end{itemize}

LLama3 exhibited exemplary performance in balancing efficiency and safety, effectively utilizing preferred zones while avoiding restricted areas. This resulted in consistent metrics across all categories, demonstrating optimal trade-offs between safety constraints and efficiency goals.

\begin{table*}[htbp]
\centering
\caption{DCIP Core Performance Metrics Across Models and Strategies}
\label{tab:pathfinding-metrics}
\scriptsize
\begin{tabular}{llrrrrrr}
\toprule
Model & Strategy & Algorithm & \begin{tabular}[c]{@{}r@{}}Nodes\\Expanded\end{tabular} & \begin{tabular}[c]{@{}r@{}}Search\\Time (s)\end{tabular} & \begin{tabular}[c]{@{}r@{}}Path\\Cost\end{tabular} & \begin{tabular}[c]{@{}r@{}}Path\\Length\end{tabular} & \begin{tabular}[c]{@{}r@{}}Number of\\Turns\end{tabular} \\
\midrule
\multicolumn{2}{l}{\textit{Baseline (avg. 10 runs)}} & A* & 2,254 & 0.1253 & 177.0 & 177 & 66 \\
\midrule
Mistral & Navigate Quickly & DCIP & 13,967 & 0.3978 & 373.0 & 179 & 9 \\
 & Maximize Safety & DCIP & 1,409 & 0.0434 & 142.0 & 177 & 26 \\
 & Balance Efficiency & DCIP & 1,442 & 0.0423 & 142.0 & 177 & 26 \\
\midrule
Llama3.1 & Navigate Quickly & DCIP & 370 & 0.0093 & 134.5 & 177 & 26 \\
 & Maximize Safety & DCIP & 1,798 & 0.0764 & 132.5 & 177 & 37 \\
 & Balance Efficiency & DCIP & 1,409 & 0.0401 & 142.0 & 177 & 26 \\
\midrule
LLama3 & Navigate Quickly & DCIP & 366 & 0.0093 & 134.5 & 177 & 26 \\
 & Maximize Safety & DCIP & 1,409 & 0.0417 & 142.0 & 177 & 26 \\
 & Balance Efficiency and Safety & DCIP & 370 & 0.0109 & 102.0 & 177 & 15 \\
\bottomrule
\end{tabular}
\end{table*}

\begin{figure}
    \centering
    \includegraphics[width=3.5in]{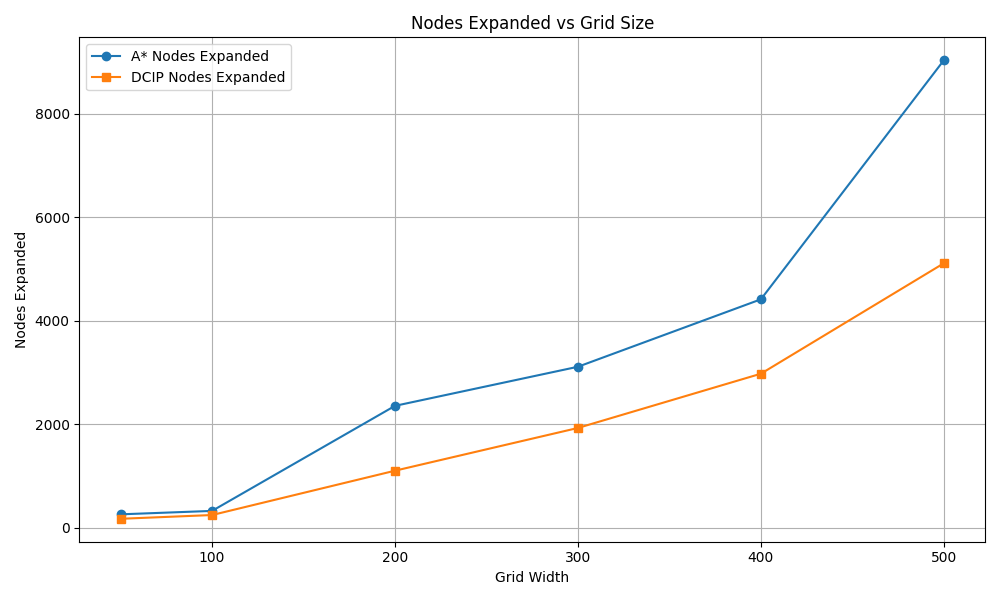}
    \caption{ The comparative analysis examines the nodes expanded between A*
and DCIP (incorporating LLAMA3.1 with Balance Efficient Strategy) across scaled environments ranging from
1 to 10 times enlargement in grid size, based on the means of 10
trials of random sampling.}
    \label{fig:enter-label1}
\end{figure}

\begin{figure}
    \centering
    \includegraphics[width=3.5in]{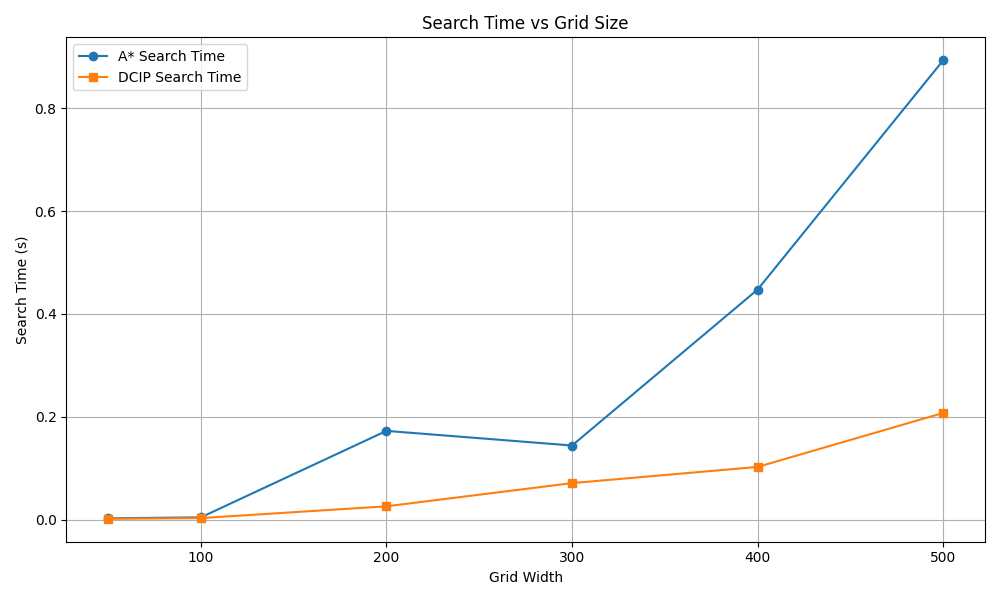}
    \caption{ The comparative analysis examines the search time between A*
and DCIP (incorporating LLAMA3.1 with Balance Efficient Strategy) across scaled environments ranging from
1 to 10 times enlargement in grid size, based on the means of 10
trials of random sampling.}
    \label{fig:enter-label2}
\end{figure}

\subsection{System Integration Testing}
To demonstrate DCIP’s versatility and effectiveness across different robotic tasks and environmental settings, we conducted system integration tests using a robot manipulator Husky mobile robot equipped with a robot arm UR3 within a simulated industrial workspace.

\subsection{Experiment Setup}
The experimental setup utilized the Husky-UR3 compound robot~\citep{chitta2017ros_control} within an industrial workspace simulated in Gazebo~\citep{koenig2004design}. The environment included shelves, storage areas, repair zones, and pedestrian agents to emulate real-world conditions. The robot was tasked with executing pick-and-place operations guided by natural language instructions, with a focus on avoiding restricted areas and ensuring safe navigation around pedestrians. This setup allowed for evaluating the DCIP framework's ability to parse instructions, update navigation plans dynamically, and adhere to environmental constraints.

\begin{figure}[t!]
\centering
	\includegraphics[width=3.3in]{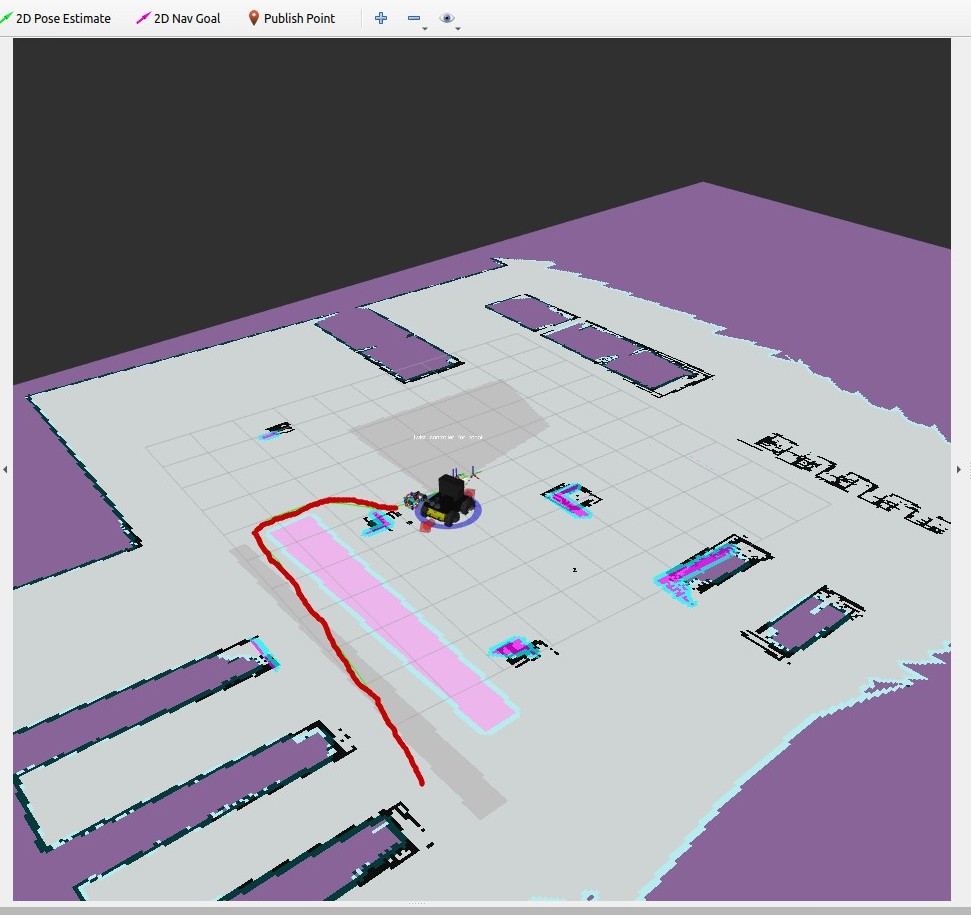}
	\caption{ Simulation experiment for navigation task in pick phase, red line representing the path planned and blocking the area(pink) considering pedestrians to navigate and prefering empty space(grey) above robot.}
	\label{fig:learning2}
	\vspace{-3mm}
\end{figure}

\subsection{Experiment Scenarios}
\begin{enumerate}
    \item \textbf{Pick Phase:}
    \begin{itemize}
        \item \textbf{Objective:} Navigate to Shelf 3 beside Self 2 to retrieve an item.
        \item \textbf{Constraints:} Avoid the repair area and maintain safe distances from pedestrians.
    \end{itemize}
    
    \item \textbf{Place Phase:}
    \begin{itemize}
        \item \textbf{Objective:} Return to the storage area to place the retrieved item.
        \item \textbf{Constraints:} Utilize empty lanes and free areas for navigation, avoid restricted zones.
    \end{itemize}
\end{enumerate}

\textbf{Procedure:}
\begin{enumerate}
    \item \textbf{Instruction Input:} Instructions generated by volunteers were fed into DCIP to obtain the corresponding action sequences.
    \item \textbf{Action Execution:} The Husky-UR3 robot executed the generated action sequences within the simulation, performing navigation and manipulation tasks as per the instructions.
    \item \textbf{Observation:} The robot’s behavior was monitored to ensure compliance with the instructions and constraints.
\end{enumerate}

%\begin{figure}[t!]
%	\includegraphics[width=3.3in]{\figpath realbot.png}
%	\caption{ In the top-left image, the robot selects an \textbf{empty area} to navigate, avoiding \textbf{pedestrian zones} for safety. The top-right image shows the mobile base choosing an walkway lane for \textbf{minimal search} effort. Finally, the bottom image illustrates the robot reaching the goal location, the printer area, following the given instruction to navigate there. }
%	\label{fig:learning}
%	\vspace{-3mm}
%\end{figure}

\begin{figure}[t!]
\centering
	\includegraphics[width=3.3in]{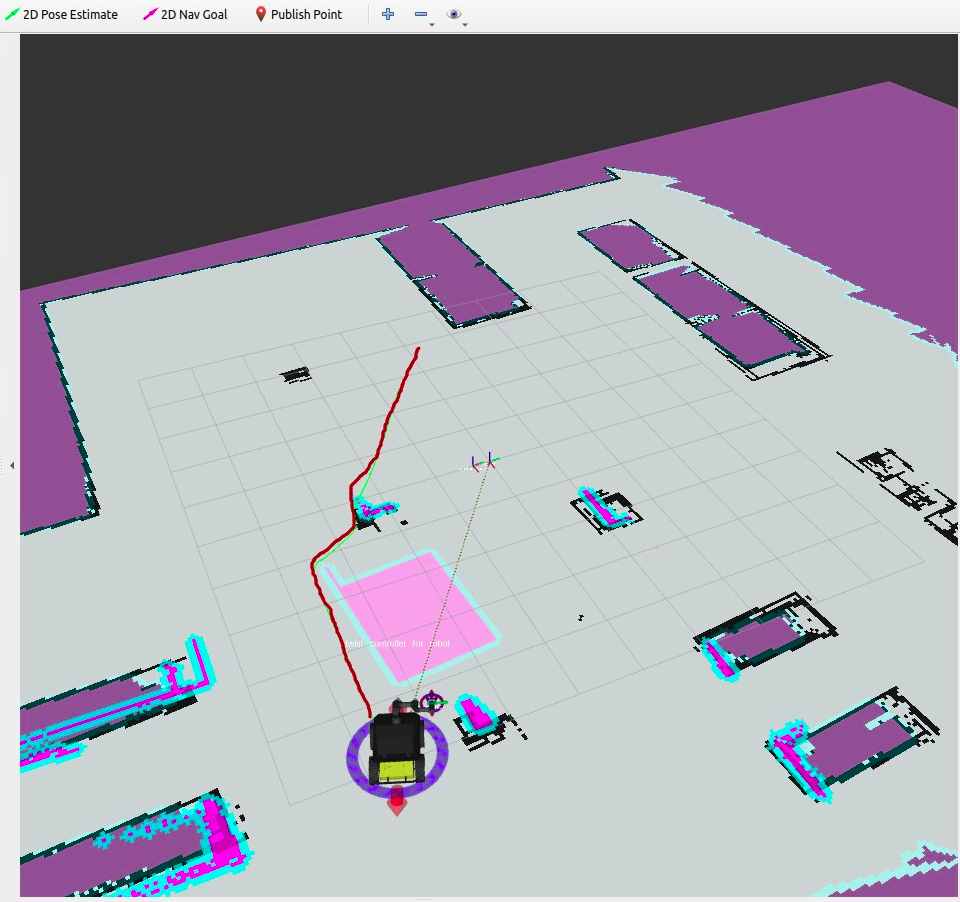}
	\caption{ Simulation experiment for navigation task in pick phase, red line representing the path planned and blocking the area (pink) considering instruction to be a repair area ahead.}
	\label{fig:learning}
	\vspace{-3mm}
\end{figure}

\subsection{Results}
The results demonstrate the effectiveness of the DCIP framework in interpreting and executing natural language instructions for autonomous navigation. The system achieved an average instruction-to-task accuracy of 90\%, with most misclassifications arising from ambiguous or multi-constraint instructions, which occasionally led to incomplete or partially incorrect actions. Despite these challenges, all pick-and-place tasks were successfully completed in the simulation, with the robot consistently complying with constraints such as avoiding repair areas and maintaining safe pedestrian distances. Additionally, DCIP dynamically updated the 2D occupancy grid to reflect preferred and restricted areas, enabling accurate and adaptive navigation planning in real-time. These outcomes highlight the robustness and reliability of DCIP in dynamic environments.
\begin{table}[!hb]
    \centering
    \caption{System Integration Performance}
    \begin{tabular}{lp{5 cm}}
        \toprule
        \textbf{Metric} & \textbf{Observations} \\
        \midrule
        Task Completion &  \\
        % \hline
        Constraint Compliance & Consistently avoided repair area and maintained pedestrian safety. \\
        % \hline
        Map Update Accuracy & Accurate updates to occupancy grid based on instructions. \\
        \bottomrule
    \end{tabular}
    \label{tab:system_integration}
\end{table}

\section{Conclusion}
This study presents the DCIP framework, a significant advancement in autonomous robotic navigation through precise parsing of natural language instructions and adaptive navigation planning. The system accurately translates complex instructions into actionable sequences and updates occupancy grids in real-time, achieving over 90\% accuracy rate in instruction parsing while ensuring safe and efficient navigation in dynamic environments. The framework successfully navigate the robot in the restricted areas and prioritizes safety, demonstrating its effectiveness in real-world collaborative scenarios. By integrating spatial understanding with adaptive planning, DCIP establishes a foundation for more intelligent and reliable autonomous robotic systems.

\bibliography{ArabRef}
\label{refrences}

\end{document}